\documentclass[11pt]{article}
\usepackage{acl2016}
\usepackage{times}
\usepackage{url}
\usepackage{latexsym}
\aclfinalcopy 
\usepackage{graphicx}

\makeatletter
\newcommand{\@BIBLABEL}{\@emptybiblabel}
\newcommand{\@emptybiblabel}[1]{}
\makeatother
\usepackage{hyperref}

\title{Flu Detector: Estimating influenza-like illness rates\\from online user-generated content}

\author{Vasileios Lampos\\
  Department of Computer Science \\
  University College London \\
  {\small\texttt{\textbf{v.lampos@ucl.ac.uk}}}\\\\
  \textbf{Flu Detector's version:} v.0.5 --- \textbf{Published on:} December 11, 2016
}

\date{}

\begin{document}
\maketitle

\begin{abstract}
We provide a brief technical description of an online platform for disease monitoring, titled as the \textbf{Flu Detector} (\href{http://fludetector.cs.ucl.ac.uk/}{\small\texttt{\textbf{fludetector.cs.ucl.ac.uk}}}). Flu Detector, in its current version (v.0.5), uses either Twitter or Google search data in conjunction with statistical Natural Language Processing models to estimate the rate of influenza-like illness in the population of England. Its back-end is a live service that collects online data, utilises modern technologies for large-scale text processing, and finally applies statistical inference models that are trained offline. The front-end visualises the various disease rate estimates. Notably, the models based on Google data achieve a high level of accuracy with respect to the most recent four flu seasons in England (2012/13 to 2015/16). This highlighted Flu Detector as having a great potential of becoming a complementary source to the domestic traditional flu surveillance schemes.
\end{abstract}

\section{Introduction}
Information epidemiology, or `infodemiology' \cite{eysenbach2009}, is evidently not a hypothesis anymore. Numerous research efforts in recent years have provided proof that user-generated data, especially in the form of search queries or social media, can be used to better understand a multi-faceted collection of health issues. Within this rapidly developing field of research, usually labelled as Computational Health, one of the most prominent examples has been the modelling of influenza-like illness (ILI) rates \cite{polgreen2008,ginsberg2009,lampos2010pandemic,culotta2010,paul2011icwsm,signorini2011}. Attempting to translate research results into an actual application, the platform of Google Flu Trends (GFT) was launched in 2008 based on a method proposed by \newcite{ginsberg2009} for mapping the frequency of search queries to ILI rates in the US. In 2010, Lampos et al. developed the first tool that used social media content to estimate ILI rates in the UK \cite{lampos2010FluDetector}. The Flu Detector of that era\footnote{Its last working snapshop (circa March 2013) is hosted under \href{http://twitter.lampos.net/epidemics/}{\texttt{twitter.lampos.net/epidemics}}} used Twitter posts and basic supervised learning models, such as the `lasso' \cite{tib1996,lampos2010pandemic} or its bootstrapped version \cite{bach2008bolasso,lampos2012}, operating on Bag-of-Words representations of the data. Naturally, there was space for further improvements, something that has been explored in various follow-up works (e.g. by \newcite{lamb2013twitter} or \newcite{Preis2014} and so on). In late 2015, amidst severe criticism \cite{olson2013,lazer2014} and bad press due to significant mispredictions in the past flu seasons, the GFT service was unfortunately discontinued.\footnote{See \href{https://www.google.org/flutrends/about/}{\texttt{google.org/flutrends}}}

Advancements in statistical Natural Language Processing (NLP) combined with a better understanding of the problem have recently led to disease models that overcome past deficiencies \cite{lampos2015intervention,lampos2015gft,yang2015pnas}. Motivated by this fact, a revamped version of \textbf{Flu Detector} (\href{http://fludetector.cs.ucl.ac.uk/}{\small\texttt{\textbf{fludetector.cs.ucl.ac.uk}}}) that has access to both Twitter and Google search data has been developed and recently launched. Given that GFT never made ILI rate estimates for England (or the UK), Flu Detector embodies the first online tool making ILI rate estimations for England based on Google search data.

\begin{figure*}[t]
    \centering
    \includegraphics[width=0.98\textwidth]{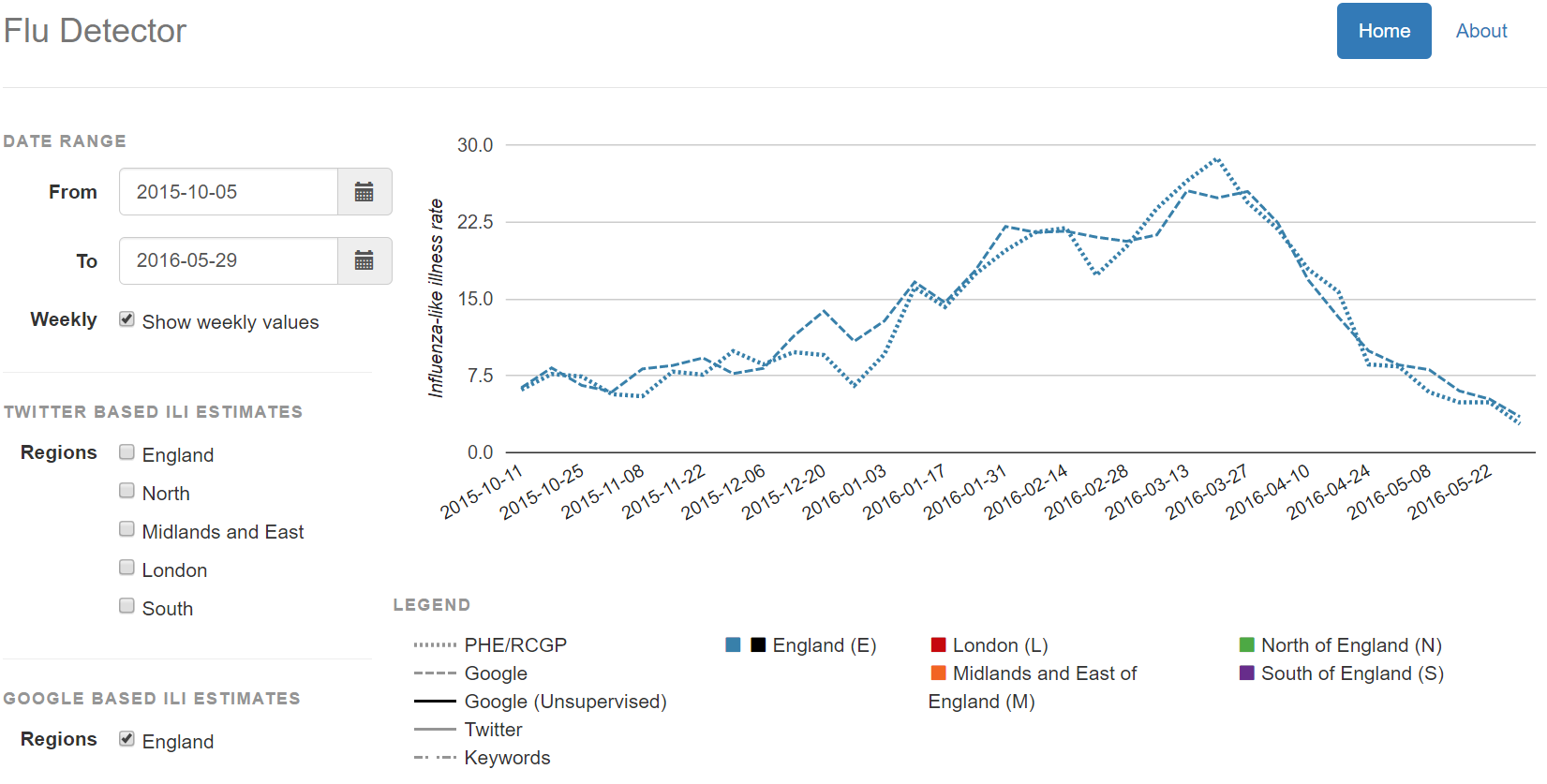}
    \caption{Flu Detector's weekly ILI estimates for the 2015/16 flu season in England based on Google search data. They are compared to the RCGP ILI rates as released by PHE.}
    \label{fig:fludetector_google_vs_rcgp}
\end{figure*}

To ensure that Flu Detector will not be a one-off scientific outcome, but will have a practical impact, the inference accuracy as well as the potential added value of the tool to the current (traditional) health surveillance schemes have been assessed\footnote{The evaluation will be published separately.} in collaboration with Public Health England (PHE), the leading governmental agency responsible for the national health surveillance schemes.\footnote{Public Health England, \href{https://www.gov.uk/government/organisations/public-health-england}{\texttt{gov.uk/government/ organisations/public-health-england}}} The results of the evaluation are positive, leading to a potential incorporation of Flu Detector's estimates as a complementary indicator in the weekly flu surveillance reports during the coming flu seasons.

This document summarises the main functionalities of Flu Detector. It should be considered as an ongoing reference to the online tool, and as such, it will be updated as new modules are being launched.

\section{Data sources}
The current version of Flu Detector has access to two online user-generated content sources, namely Twitter and Google search. The supervised models of ILI for England are trained based on syndromic surveillance data.

\subsection{Twitter}
We collect approximately every exactly geolocated tweet in England using Twitter's Streaming API.\footnote{Twitter Streaming API, \href{https://dev.twitter.com/streaming/overview}{\texttt{dev.twitter.com/ streaming/overview}}} By ``exact geolocation'' we refer to tweets, where the geo-coordinates (latitude and longitude) of the user who posted them, are available. This amounts to an average of approximately $350{,}000$ tweets per day. We note that this number is relatively small as, according to our estimates, it represents only $1\%$-$2\%$ of the entire set of tweets from users in England. Hence, the ILI rate inferences based on Twitter data are inevitably unstable.

\subsection{Google search}
Flu Detector has access to a non standardised version of the publicly available Google Trends outputs through a private Google Health Trends API.\footnote{The Google Health Trends API can only be used for academic research with a health-oriented focus.} This provides (aggregate and anonymised) normalised frequencies of search queries. More specifically, a query frequency expresses the probability of a short search session for a specific geographical region and temporal resolution, drawn from a uniformly distributed $10\%$-$15\%$ sample of all corresponding search sessions.

\subsection{Syndromic surveillance}
At the moment, Flu Detector models ILI rates as reported by the Royal College of General Practitioners (RCGP) and PHE.\footnote{See \href{https://www.gov.uk/government/statistics/weekly-national-flu-reports}{\texttt{gov.uk/government/statistics/weekly -national-flu-reports}}} The estimates represent the number of doctor consultations reporting ILI symptoms per $100{,}000$ people in England.

\section{Statistical models and performance evaluation}
Supervised learning techniques are used to model flu rates from Twitter or Google search data. A selection of papers has served as motivation for the actual methods that are employed on the website, from early papers on the topic \cite{ginsberg2009,lampos2010pandemic} to most recent developments \cite{lampos2015intervention,lampos2015gft,zou2016iid}. The applied methods combine these different pieces of knowledge with advancements in statistical NLP (e.g. the use of neural word embeddings \cite{mikolov2013iclr,mikolov2013word2vec}) and, at the moment, are being documented. 

As a preliminary performance indicator of the Google search based model, the average Mean Absolute Error in year-long weekly ILI rate estimates across four flu seasons (from 2012/13 to 2015/16) is approximately equal to $1.5$ (in $100{,}000$ people) compared to the corresponding RCGP ILI rates; the corresponding average Pearson correlation is equal to $.95$. Extensive performance evaluation will become available in forthcoming publications.

\section{Back-end and front-end operations}
At the back-end of Flu Detector, there is a software pipeline for data collection, storage and processing. The latter uses standard Python libraries (e.g. {\small\texttt{gensim}},
{\small\texttt{nltk}}, {\small\texttt{numpy}}, {\small\texttt{scipy}} and so on) and the Apache Hadoop framework\footnote{Apache Hadoop, \href{http://hadoop.apache.org/}{\texttt{hadoop.apache.org}}} for task parallelisation. Textual data can be manually processed in batches (e.g. for model training). In addition, the frequency of the textual variables used in Flu Detector's models is being automatically updated on a daily basis.

The ILI estimation models, which are trained offline, are used to produce daily (over-night) inferences as well as weekly ones. To maintain a consistency with the data distributions during the model training phase, where only weekly ILI rates are available, each estimate on Flu Detector (even the daily ones) uses a week-long set of observations. For example, to estimate the ILI rate of date $i$, we use the frequencies of textual terms during the dates $\{i,i-1,\dots,i-6\}$ for the target data set. For Twitter-driven estimates, which are consequently based on a small portion of data, and tend to be noisy, the user of the website can also access smoothed versions of the inferred time series.

The current version of Flu Detector incorporates $5$ Twitter-based models, one focusing on England as a whole and the rest in sub-regions (`London', `North England', `South England', `Midlands and East England'). As expected, the regional models are very unstable given the even smaller data ratio that characterises them. Moreover, the platform has a Google search model for England only (regional Google search data have not yet been made available). Given the higher penetration of Google search in the real population as well as the significantly larger sample of searches that are used to compute search query frequencies ($10\%$-$15\%$), the corresponding estimates are much more reliable.

Apart from its public interface, Flu Detector has also an internal one, used for testing new modules and evaluating estimates compared to traditional syndromic surveillance schemes (see Fig.~\ref{fig:fludetector_google_vs_rcgp}).

\section{Conclusions and future work}
In this brief report, we introduced Flu Detector, an online tool for presenting disease rate estimates based on user-generated content. The current version of Flu Detector uses data from Google search or Twitter and displays ILI rate estimates for England. This report will be updated as new functionalities are being launched. 

Future work includes the consideration of different infectious diseases, the incorporation of more data sources as well as the development of unsupervised disease modelling schemes. Stratified disease estimates based on perceived user demographics, e.g. age \cite{rao2010}, occupation or socioeconomic status \cite{preotiuc2015jobs,preotiuc2015plos,lampos2016ses}, as well as the expansion of models so as to cover different countries are among our priorities.

\paragraph{Acknowledgements} Flu Detector is funded by the EPSRC project EP/K031953/1 (or i-sense)\footnote{EPSRC IRC project i-sense, \href{https://www.i-sense.org.uk/}{\texttt{i-sense.org.uk}}} and by a Google Research sponsorship. V. Lampos would like to thank all the people involved in the various stages of development of Flu Detector and the underlying methods, and in particular I.J. Cox, A.C. Miller, J.K. Geyti, B. Zou, M. Wagner and R. Pebody. He would also like to thank PHE, the RCGP and Google for providing data. Credit should also be given to N. Cristianini and T. De Bie who participated in the development of Flu Detector's predecessor \cite{lampos2010FluDetector}.

{
\small
\bibliography{refs}

\begin{thebibliography}{}

\bibitem[\protect\citename{Bach}2008]{bach2008bolasso}
Francis~R. Bach.
\newblock 2008.
\newblock {Bolasso: Model Consistent Lasso Estimation Through the Bootstrap}.
\newblock In {\em Proc. of the 25th International Conference on Machine
  Learning}, pages 33--40.

\bibitem[\protect\citename{Culotta}2010]{culotta2010}
Aron Culotta.
\newblock 2010.
\newblock {Towards Detecting Influenza Epidemics by Analyzing Twitter
  Messages}.
\newblock In {\em Proc. of the 1st Workshop on Social Media Analytics}, pages
  115--122.

\bibitem[\protect\citename{Eysenbach}2009]{eysenbach2009}
Gunther Eysenbach.
\newblock 2009.
\newblock {Infodemiology and Infoveillance: Framework for an Emerging Set of
  Public Health Informatics Methods to Analyze Search, Communication and
  Publication Behavior on the Internet}.
\newblock {\em Journal of Medical Internet Research}, 11(1):e11.

\bibitem[\protect\citename{Ginsberg \bgroup et al.\egroup }2009]{ginsberg2009}
Jeremy Ginsberg, Matthew~H. Mohebbi, Rajan~S. Patel, et~al.
\newblock 2009.
\newblock {Detecting influenza epidemics using search engine query data}.
\newblock {\em Nature}, 457(7232):1012--1014.

\bibitem[\protect\citename{Lamb \bgroup et al.\egroup }2013]{lamb2013twitter}
Alex Lamb, Michael~J. Paul, and Mark Dredze.
\newblock 2013.
\newblock {Separating Fact from Fear: Tracking Flu Infections on Twitter}.
\newblock In {\em Proc. of the 2013 Conference of the North American Chapter of
  the Association for Computational Linguistics: HLT}, pages 789--795.

\bibitem[\protect\citename{Lampos and Cristianini}2010]{lampos2010pandemic}
Vasileios Lampos and Nello Cristianini.
\newblock 2010.
\newblock {Tracking the flu pandemic by monitoring the Social Web}.
\newblock In {\em Proc. of the 2nd International Workshop on Cognitive
  Information Processing}, pages 411--416.

\bibitem[\protect\citename{Lampos and Cristianini}2012]{lampos2012}
Vasileios Lampos and Nello Cristianini.
\newblock 2012.
\newblock {Nowcasting Events from the Social Web with Statistical Learning}.
\newblock {\em ACM Transactions on Intelligent Systems and Technology},
  3(4):1--22.

\bibitem[\protect\citename{Lampos \bgroup et al.\egroup
  }2010]{lampos2010FluDetector}
Vasileios Lampos, Tijl De~Bie, and Nello Cristianini.
\newblock 2010.
\newblock {Flu Detector: Tracking Epidemics on Twitter}.
\newblock In {\em Proc. of the 2010 European Conference on Machine Learning and
  Knowledge Discovery in Databases}, pages 599--602.

\bibitem[\protect\citename{Lampos \bgroup et al.\egroup }2015a]{lampos2015gft}
Vasileios Lampos, Andrew~C. Miller, Steve Crossan, and Christian Stefansen.
\newblock 2015a.
\newblock {Advances in nowcasting influenza-like illness rates using search
  query logs}.
\newblock {\em Scientific Reports}, 5(12760).

\bibitem[\protect\citename{Lampos \bgroup et al.\egroup
  }2015b]{lampos2015intervention}
Vasileios Lampos, Elad Yom-Tov, Richard Pebody, and Ingemar~J. Cox.
\newblock 2015b.
\newblock {Assessing the impact of a health intervention via user-generated
  Internet content}.
\newblock {\em Data Mining and Knowledge Discovery}, 29(5):1434--1457.

\bibitem[\protect\citename{Lampos \bgroup et al.\egroup }2016]{lampos2016ses}
Vasileios Lampos, Nikolaos Aletras, Jens~K. Geyti, Bin Zou, and Ingemar~J. Cox.
\newblock 2016.
\newblock {Inferring the Socioeconomic Status of Social Media Users Based on
  Behaviour and Language}.
\newblock In {\em Proc. of 38th European Conference on IR Research}, pages
  689--695.

\bibitem[\protect\citename{Lazer \bgroup et al.\egroup }2014]{lazer2014}
David Lazer, Ryan Kennedy, Gary King, and Alessandro Vespignani.
\newblock 2014.
\newblock {The Parable of Google Flu: Traps in Big Data Analysis}.
\newblock {\em Science}, 343(6176):1203--1205.

\bibitem[\protect\citename{Mikolov \bgroup et al.\egroup
  }2013a]{mikolov2013iclr}
Tomas Mikolov, Kai Chen, Greg~S. Corrado, and Jeff Dean.
\newblock 2013a.
\newblock {Efficient Estimation of Word Representations in Vector Space}.
\newblock In {\em Proc. of the International Conference on Learning
  Representations, Workshop Track}, pages 1--12.

\bibitem[\protect\citename{Mikolov \bgroup et al.\egroup
  }2013b]{mikolov2013word2vec}
Tomas Mikolov, Ilya Sutskever, Kai Chen, Greg~S. Corrado, and Jeff Dean.
\newblock 2013b.
\newblock {Distributed Representations of Words and Phrases and their
  Compositionality}.
\newblock In {\em Advances in Neural Information Processing Systems 26}, pages
  3111--3119.

\bibitem[\protect\citename{Olson \bgroup et al.\egroup }2013]{olson2013}
Donald~R. Olson, Kevin~J. Konty, Marc Paladini, Cecile Viboud, and Lone
  Simonsen.
\newblock 2013.
\newblock {Reassessing Google Flu Trends Data for Detection of Seasonal and
  Pandemic Influenza: A Comparative Epidemiological Study at Three Geographic
  Scales}.
\newblock {\em PLOS Computational Biology}, 9(10), 10.

\bibitem[\protect\citename{Paul and Dredze}2011]{paul2011icwsm}
Michael~J. Paul and Mark Dredze.
\newblock 2011.
\newblock {You Are What You Tweet: Analyzing Twitter for Public Health}.
\newblock In {\em Proc. of the 5th International Conference on Weblogs and
  Social Media}, pages 265--272.

\bibitem[\protect\citename{Polgreen \bgroup et al.\egroup }2008]{polgreen2008}
Philip~M. Polgreen, Yiling Chen, David~M. Pennock, Forrest~D. Nelson, and
  Robert~A. Weinstein.
\newblock 2008.
\newblock {Using Internet Searches for Influenza Surveillance}.
\newblock {\em Clinical Infectious Diseases}, 47(11):1443--1448.

\bibitem[\protect\citename{Preis and Moat}2014]{Preis2014}
Tobias Preis and Helen~Susannah Moat.
\newblock 2014.
\newblock {Adaptive nowcasting of influenza outbreaks using~Google searches}.
\newblock {\em Open Science}, 1(2).

\bibitem[\protect\citename{Preo\c{t}iuc-Pietro \bgroup et al.\egroup
  }2015a]{preotiuc2015jobs}
Daniel Preo\c{t}iuc-Pietro, Vasileios Lampos, and Nikolaos Aletras.
\newblock 2015a.
\newblock {An analysis of the user occupational class through Twitter content}.
\newblock In {\em Proc. of the 53rd Annual Meeting of the Association for
  Computational Linguistics (Volume 1: Long Papers)}, pages 1754--1764.

\bibitem[\protect\citename{Preo\c{t}iuc-Pietro \bgroup et al.\egroup
  }2015b]{preotiuc2015plos}
Daniel Preo\c{t}iuc-Pietro, Svitlana Volkova, Vasileios Lampos, Yoram Bachrach,
  and Nikolaos Aletras.
\newblock 2015b.
\newblock {Studying User Income through Language, Behaviour and Affect in
  Social Media}.
\newblock {\em PLOS ONE}, 10(9).

\bibitem[\protect\citename{Rao \bgroup et al.\egroup }2010]{rao2010}
Delip Rao, David Yarowsky, Abhishek Shreevats, and Manaswi Gupta.
\newblock 2010.
\newblock {Classifying Latent User Attributes in Twitter}.
\newblock In {\em Proc. of the 2nd International Workshop on Search and Mining
  User-generated Contents}, pages 37--44.

\bibitem[\protect\citename{Signorini \bgroup et al.\egroup
  }2011]{signorini2011}
Alessio Signorini, Alberto~Maria Segre, and Philip~M. Polgreen.
\newblock 2011.
\newblock {The Use of Twitter to Track Levels of Disease Activity and Public
  Concern in the U.S. during the Influenza A H1N1 Pandemic}.
\newblock {\em PLOS ONE}, 6(5).

\bibitem[\protect\citename{Tibshirani}1996]{tib1996}
Robert Tibshirani.
\newblock 1996.
\newblock {Regression Shrinkage and Selection via the Lasso}.
\newblock {\em Journal of the Royal Statistical Society: Series B (Statistical
  Methodology)}, 58(1):267--288.

\bibitem[\protect\citename{Yang \bgroup et al.\egroup }2015]{yang2015pnas}
Shihao Yang, Mauricio Santillana, and Samuel~C. Kou.
\newblock 2015.
\newblock {Accurate estimation of influenza epidemics using Google search data
  via ARGO}.
\newblock {\em Proceedings of the National Academy of Sciences},
  112(47):14473--14478.

\bibitem[\protect\citename{Zou \bgroup et al.\egroup }2016]{zou2016iid}
Bin Zou, Vasileios Lampos, Russell Gorton, and Ingemar~J. Cox.
\newblock 2016.
\newblock {On Infectious Intestinal Disease Surveillance Using Social Media
  Content}.
\newblock In {\em Proc. of the 6th International Conference on Digital Health
  Conference}, pages 157--161.

\end{thebibliography}
\bibliographystyle{acl2016}
}
\end{document}